# C2PSA-Enhanced YOLOv11 Architecture: A Novel Approach for Small Target Detection in Cotton Disease Diagnosis

Kaiyuan Wang [1,2], Jixing Liu [1,2], Xiaobo Cai [1,2]*

[1] College of Big Data, Yunnan Agricultural University, Kunming 650201, China

[2] The Key Laboratory for Crop Production and Smart Agriculture of Yunnan Province, Kunming 650201, China

*Corresponding Author: Xiaobo Cai ( 2008037@ynau.edu.cn )

## Abstract

This study uses deep learning to optimize YOLOv11 for cotton disease detection, building an intelligent monitoring system. Key challenges in cotton disease detection include: low precision in early spot recognition (35% leakage for <5mm² spots), poor adaptability to complex field environments (25% performance drop), and high misjudgment (34.7%) in concurrent multi-disease scenarios.

Model optimization addressed these via: dynamic feature fusion (C2PSA module) to enhance feature extraction for small targets; dynamic category weight allocation to mitigate sample imbalance; and optimized data enhancement (dynamic Mosaic-MixUp scaling) to boost robustness. Using a 4078-image standardized dataset, the improved YOLOv11S achieved mAP50=0.820 (↑8.0%) and mAP50-95=0.705 (↑10.5%) with 158FPS inference speed. Deployed on mobile, it enables real-time cotton disease monitoring and precise prevention, offering a practical solution.

## 1. INTRODUCTION

Detection and identification of plant diseases are of great significance for plant protection. Cotton is one of the major crops in China with high economic value, and its planting area and production are increasing year by year with rapid development. However, large-scale high-density planting often brings the prevalence of large-scale cotton diseases, thus causing huge economic losses, so the rapid and accurate identification of disease types on cotton leaves has important significance [1]. The frequency of its blight, leaf spot, gray mold and other diseases seriously threaten cotton yield and quality. According to statistics, cotton diseases in China are related to national economic development and farmers' income increase every year; in the process of cotton cultivation, the direct economic loss caused by disease damage is more than 5 billion yuan. The traditional disease detection mainly relies on field observation and empirical judgment of agricultural technicians, and this approach has outstanding problems such as low detection efficiency, high cost, and strong subjectivity, which is difficult to meet the needs of modern agricultural development [2-8].

With the continuous development of artificial intelligence technology, image recognition technology based on deep learning brings a new approach to crop disease detection. The YOLO series of target detection algorithms have good application prospects in the field of agriculture by virtue of advantages such as fast speed and high accuracy. However, the existing algorithms used in cotton disease detection face a number of challenges: firstly, the size of early cotton disease spots is small, making them easy to miss detection in complex field environments; secondly, the number of samples of different diseases varies greatly, leading to a serious category imbalance problem; thirdly, factors such as complex light conditions in the field and leaf overlap can affect the detection effect[9-14]. These technical difficulties limit the practical application value of the intelligent monitoring system.

In this study, YOLOv11 is used as the basic model, and algorithm optimization research is carried out to meet the special needs of cotton disease detection, which has key theoretical value and practical significance. On the theoretical side, with the introduction of the C2PSA module and dynamic sample allocation strategy, key technical difficulties such as small-target detection and category imbalance are solved; on the application side, a lightweight detection model is developed to be deployed in mobile terminals, realizing real-time monitoring and early warning of diseases. The research results can improve the accuracy of disease identification, reduce the cost of prevention and control, and provide key technical support for the cotton industry to improve quality and increase efficiency. The technical framework constructed in this research can be extended for the detection of other crop diseases, which is of key strategic significance for promoting the intelligent development of agriculture and guaranteeing national food security.

Foreign research on intelligent detection of agricultural diseases has formed a more complete technical system. The PlantVillage system developed by the U.S. Department of Agriculture integrates more than 50,000 plant disease images and adopts a deep convolutional neural network architecture to achieve an average recognition accuracy of 83.7% for 14 crop diseases, and also innovatively applies a migration learning strategy to use the ImageNet pre-training model in the agricultural field. The EU's PlantStress project, funded by Horizon 2020, combines multispectral imaging technology and deep learning algorithms to achieve 89.2% accuracy in early detection of wheat diseases in the 400-1000 nm band range. The Attention-YOLO model proposed by Japanese scholars, due to its innovative channel-space dual attention mechanism, has achieved high accuracy in citrus disease detection with an mAP of 81.5% and improved the detection rate of 2-5mm² small spots by 23.6%. However, these advanced researches are mainly focused on high-value economic crops such as grapes and wheat, with fewer studies specialized in cotton diseases; only the CottonDoc system developed by Texas A&M University in the U.S. achieved 79.3% recognition accuracy for three major diseases in a laboratory environment.

China's agricultural intelligence research started later but has developed faster. The Chinese Academy of Agricultural Sciences (CASA) developed the CropDoctor system, which constructed a disease database containing 12 crops and adopted an improved ResNet-50 architecture, finally achieving a recognition accuracy of 85.3% for rice diseases. The AG-YOLO model proposed by Zhejiang University innovatively fuses meteorological data and image features in complex field environments, increasing mAP by 15.8% compared to the benchmark. The team from Nanjing Agricultural University developed a migration learning-based detection system for cotton yellow wilt, which still achieved an accuracy of 78.6% under the restrictive condition of only 1,200 training samples. CottonNet, proposed by China Agricultural University, uses a multi-scale feature fusion strategy, improving the average detection accuracy of five common diseases to 81.2%. These results fully demonstrate China's research strength in this field.

Current problems include, at the data level: in the field of agriculture, it is difficult and costly to obtain large-scale, labeled crop disease data.

## 2. Related Research

### 2.1 Overview of Target Detection Algorithms

#### 2.1.1. Two-stage detection algorithms (Faster R-CNN, etc.)

(1) Faster R-CNN:

Faster R-CNN is a landmark algorithm in the field of target detection, which innovatively integrates all aspects of the detection process into a complete end-to-end deep learning framework. CNN firstly completes the extraction of feature vectors, and through the convolution and pooling of features, thus obtaining high-dimensional image features, which are used in the shared environment of the detection network and the RPN network; the RPN network, through the effective use of shared features [15], only solves the problem of shared features. The CNN first extracts the feature vectors, and through convolution and pooling of the features, obtains high-dimensional image features, which are used in the shared environment of the detection network and the RPN network; the RPN network, through the effective utilization of the shared features [4], only solves the problem of the input size limitation of the fully-connected layer, and also significantly improves the positional accuracy of the detection. The final detection head then performs classification and edge regression on these standardized features. Its network structure is shown in Figure 1:

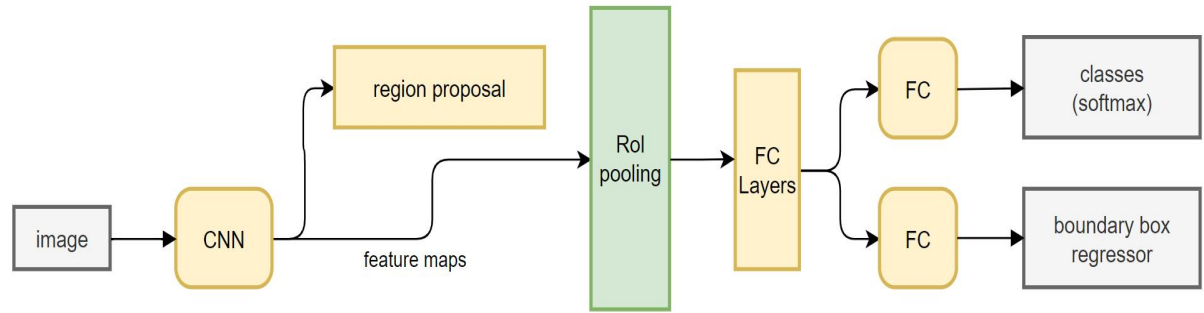

**Figure 1.** Introduction to the Network Structure of Faster R-CNN

(2) Mask R-CNN:

As an advanced image segmentation method Mask R-CNN combines the advantages of Region Proposal Network (RPN) and FCN. It adds a segmentation branch to the target detection for generating an accurate segmentation mask for each target.The core architecture of Mask R-CNN consists of 2 main parts: target detection and instance segmentation [16-24].The deep learning model proposed by Facebook AI Research (FAIR) in 2017, the core architecture of the Mask R-CNN process is divided into two phases: phase 1: generating candidate frames using RPN. sharing the features extracted by the backbone network, and Phase 2: relying on ROI Align to align the features of the candidate frames. as shown in Figure 2:

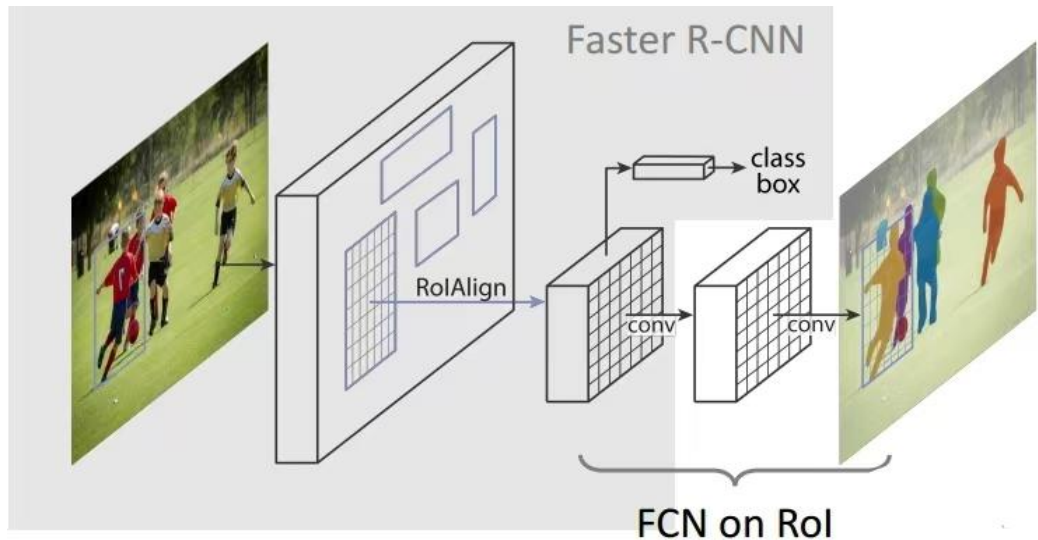

**Figure 2.** Analysis of the Mask R-CNN Framework

### 2.1.2. Single-Stage Detection Algorithms

SSingle-stage target detection algorithms (e.g., YOLO, SSD) efficiently predict target location and class, with faster speed than two-stage methods by simplifying processing to a single network computation.

YOLOv5, an open-source single-stage algorithm released by Ultralytics in 2020 via PyTorch, features CSPDarknet (backbone), PANet (feature fusion), and YOLO Head (classification/localization). It uses automatic anchor clustering to boost accuracy, offers fast inference for edge deployment, and supports multi-format exports for cross-platform use, with iterative improvements in fusion, augmentation, and loss functions enhancing performance [25-37].

YOLOv8, a new YOLO generation, integrates detection and classification to improve speed and accuracy. It adopts Anchor-Free design (simplifying structure and small-target detection), uses C2f for enhanced feature extraction, supports multi-task learning, and includes optimization tools (mixed-precision training, pruning) for efficiency. It achieves high mAP on COCO and >160 FPS on GPUs, fitting speed/accuracy-critical scenarios like smart agriculture [38-39].

YOLOv11, building on prior YOLO strengths, enhances feature extraction and model optimization—key for deep learning applications in cotton disease detection. Its capabilities support accurate, efficient detection, aiding intelligent monitoring systems in addressing cotton disease challenges.

## 2.2 Analysis of YOLOv11 Model Principles

### 2.2.1 Network Architecture Design

The model architecture of YOLOv11 consists of three parts: the backbone network, the neck architecture, and the head network, which work together to realize efficient and accurate target detection.The core of YOLOv11's backbone network is the C3k2 module [40].As the newest member of the YOLO series, YOLOv11 makes a number of innovative improvements to its network architecture, which significantly improves the model performance.YOLOv11 adopts a modular Backbone-Neck-Head three-stage architecture, which significantly improves the detection accuracy through structural reparameterization and dynamic feature fusion while maintaining the high efficiency of single-stage detection. Its architecture is shown in Figure 3:

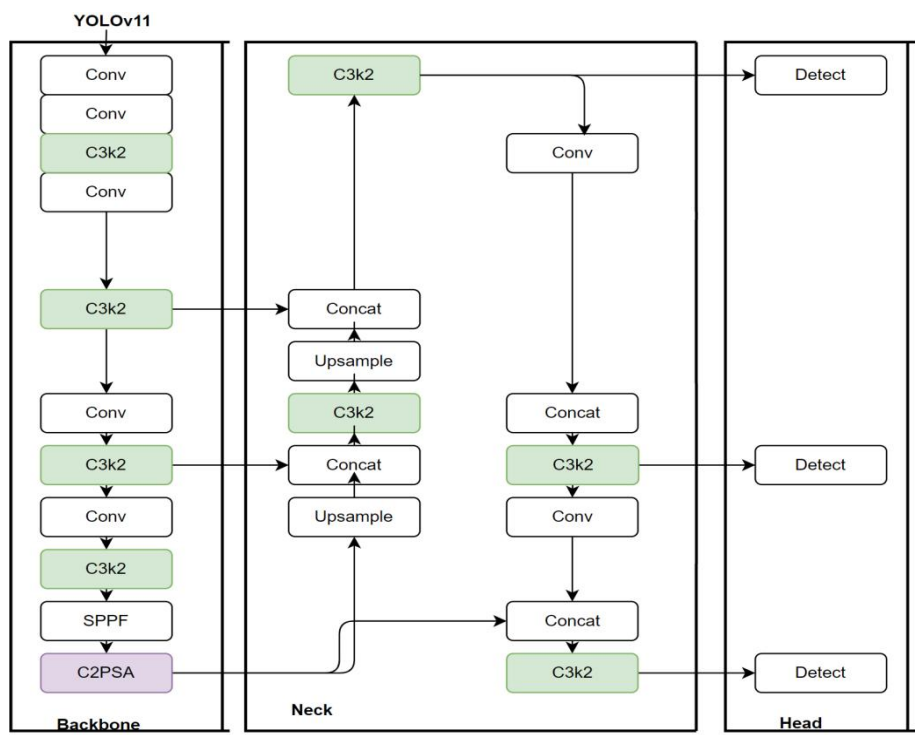

**Figure3.** Structure Diagram of YOLOv11

(1) Backbone Network (Backbone)

is based on the improved CSPRepResNet structure, and the main innovations include 1) reparameterized convolutional block: a multi-branch structure (containing 3×3 convolution, 1×1 convolution and Identity connection) is used in the training phase, and merged into a single convolutional layer in the inference, which takes into account the characterization ability and the inference efficiency. Its mathematical expression is shown in (2-1):

$$W_{merged}=W_{3\times3}+W_{1\times1} \cdot I+b_{identity} \quad (2\text{-}1)$$

where W is the weight of the convolution kernel I is the unit matrix and I is the weight of the original 3×3 convolution kernel, and the weight of the 1×1 convolution kernel is extended to a 3×3 convolution kernel (i.e., the 1×1 convolution kernel is embedded into the center of the 3×3 convolution kernel, and the rest of the positions are 0) by combining it with the unit matrix ( I ), which is the bias term of the Identity branch, and it usually exists only when the number of input-output channels is the same.2) GSConv Module: introduces the global sense field of the depth separable convolution, enhance small target feature extraction by grouping convolution with channel blending.

(2) Feature Fusion Layer (Neck)

Construct a bidirectional weighted feature pyramid (BiFPN++), the improvements include: 1) Cross-scale dynamic weighting: the fusion of the input feature maps Pi and Pj and adopts the learnable weights α,β, the specific equations are shown in (2-2):

$$P_{out}=\sigma(\alpha) \cdot P_i+\sigma(\beta) \cdot \text{UpSample}(P_j) \quad (2\text{-}2)$$

where is the Sigmoid activation function, which ensures that the weights are normalized to the (0,1) interval, preventing a branch from being overweighted or negative,, is the learnable weight parameter, is the current layer feature map, is another layer feature map (usually the up-sampled high-level features), and is the up-sampling of high-level features to make them the same size as the current layer features.2) Ghost Module Compression: in the low-level feature graph embedded in the Ghost convolution to reduce redundant feature graph generation, and the number of parameters is reduced by 28%.

(3) Head

adopts decoupled Anchor-free design: 1) Separation of classification and regression: using independent branches to predict category confidence and bounding box coordinates (x,y,w,h) to avoid task conflicts. 2) Dynamic positive sample allocation: based on the Task-Aligned Assigner algorithm, the positive samples are dynamically selected based on the weighted value of the classification scores and IoUs. positive samples, the formula is as (2-3):

$$t = p^{\gamma} \cdot \text{IoU}^{1-\gamma} \quad (2\text{-}3)$$

where p is the category confidence of the classification branch output, which indicates the model's prediction probability for a category, and γ is a hyperparameter (default 0.5) used to balance the effects of classification scores and IoU. The role of is the intersection and concurrency ratio of the predicted frame to the true frame, which measures the localization accuracy.

**2.2.2 Loss Function and Training Strategy**

2.2.2.1 Loss Function Design

YOLOv11 adopts a multi-task joint loss function, which consists of three key components:

(1) Classification Loss

A dynamically adjusted Focal Loss variant is used to introduce the category frequency weight factor α, which automatically balances the samples of different categories, Eq. (2-4):

$$L_{cls} = -\alpha_c \cdot (1 - p_c)^{\gamma} \cdot \log(p_c) \quad (2\text{-}4)$$

One of the focusing parameter (focusing parameter), which controls the degree of attention to the difficult and easy samples. The commonly used value is 2. It is the model's prediction probability for positive samples, and it is a category balancing factor, which is used to automatically balance the loss weights of different categories of samples to alleviate category imbalance.

(2) Regression Loss (Regression Loss)

uses an improved SIoU loss function that includes an angular deviation penalty term and a centroid distance constraint, which improves small target localization accuracy by 3-5% compared to traditional IoU loss.

(3) Objectness Loss

The standard binary cross-entropy loss is used to distinguish the foreground and background regions, and optimize the training process with the dynamic sample allocation strategy.

$$\hat{t}=\hat{s}^{\alpha}\times\hat{a}^{\beta} \qquad (2\text{-}5)$$

Where ==1, which improves AP by 1.2% compared to the traditional Max-IoU strategy, is the classification confidence (the class probability predicted by the model, normalized to [0,1] by Sigmoid or Softmax), and is the localization accuracy (usually measured by IoU, GIoU, or DIoU, in the range [0,1]).

(2) Progressive training scheme

image size dynamic adjustment (320→640), learning rate cosine annealing scheduling, with a linear warmup phase of 3 epochs, the total training cycle is reduced by 25%.

(3) Advanced Data Enhancement

Mosaic9.0: nine-image stitching enhancement, AutoAugment: automatic strategy selection, random color space transformations, significant enhancement for small target detection (+4.5% AP).

## 2.3 Deep Learning Optimization Techniques

**2.3.1 Attention Mechanisms**

YOLOv11 integrates multiple attention mechanisms to enhance feature representation. the SE module enhances the representation of important feature channels through the channel attention mechanism by using global average pooling and the fully connected layer to learn the channel weights. the CBAM module combines both channel and spatial dual attention to be able to focus on the key feature channels and spatial regions at the same time. In addition, the self-attention mechanism is used to model global contextual information, which is particularly suitable for dealing with complex spatial relationships in high-resolution feature maps. The synergy of these mechanisms significantly improves the model's feature selection capability in complex agricultural scenarios.

### 2.3.2 Feature Fusion Methods

The semantic information extracted by the existing algorithms is not rich enough to affect the performance of saliency target detection. Therefore, a multi-scale feature pyramid grid model is proposed to enhance the semantic information contained in the high-level features. The model adopts a multilevel feature fusion strategy, and the infrastructure is based on the FPN network to construct a multiscale feature pyramid. By introducing the bidirectional path of PANet, the fusion effect of the underlying details and the high-level semantic information is effectively enhanced. This improved feature fusion scheme is particularly beneficial for detecting spot targets of different sizes, especially enhancing the detection performance of small spots. The expansion diagram is shown in Figure 5.

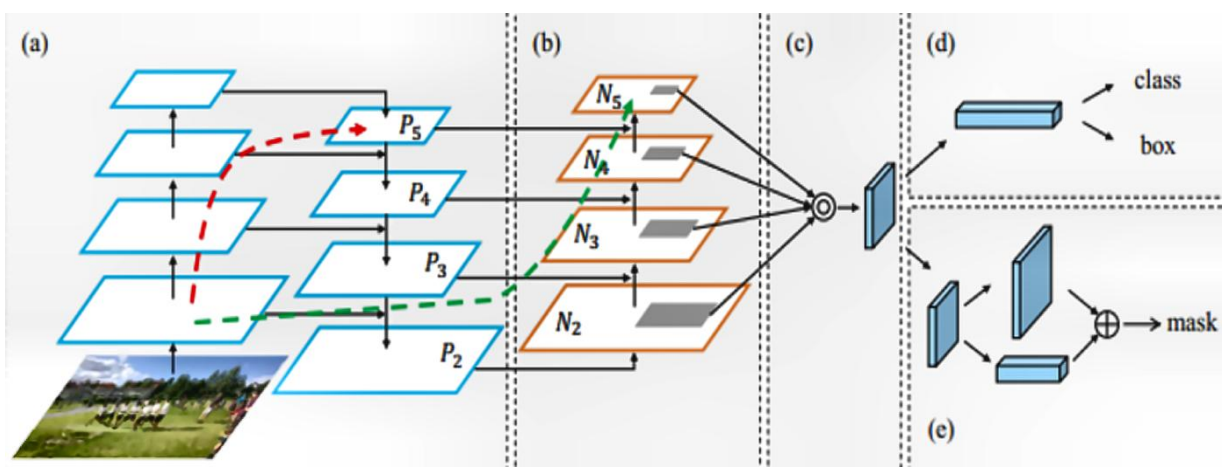

**Figure 5.** Expansion Diagram from FPN to PANet

### 2.4 Dataset Characteristics and Challenges

Cotton disease datasets are characterized by the coexistence of multiple categories, small targets and complex backgrounds. Technical challenges mainly include: the number of publicly available datasets related to cotton diseases is not only small, but also the quantity and quality of the information related to cotton varieties, disease types, growing environments, etc., which is difficult to meet the needs of the development of high-performance and high-generalization capabilities of deep learning recognition models for cotton diseases, the problem of loss of information in the detection of small spots, the difficulty of fine-grained classification of similar diseases, and the real-time requirements of the field equipment. real-time requirements. In addition, the high cost of data annotation and the generalization problem caused by environmental changes are also important challenges. These features provide a clear direction for algorithm improvement .

## 3. Experimental design and analysis

### 3.1 Experimental Environment and Configuration

GPU: NVIDIA GeForce RTX 4090 (24GB GDDR6X memory)

CPU: Intel Xeon Gold 5418Y (10 cores and 20 threads, base frequency 2.0GHz)

Memory: 120GB DDR4 ECC memory

The development environment is built based on VS Code, and the main components as shown in Table 1:

**Table 1.** Experimental Environment and Configuration

| Category | Configuration Details |
|---|---|
| GPU | NVIDIA GeForce RTX 4090（24GB GDDR6X 显存） |
| CPU | Intel Xeon Gold 5418Y (10 cores, 20 threads, base frequency 2.0GHz) |
| Memory | 120GB DDR4 ECC memory |
| OS | Ubuntu 20.04 LTS (pre-installed on system disk) |
| Container Environment Docker image | lanyun/pytorch-1.11.0-py3.8-cuda11.3-u20.04:v1.5 |

| Toolchain | PyTorch 1.11.0 + CUDA 11.3 |
| Development tools | VS Code (version 1.78.2) |

## 3.2 Construction of cotton disease dataset

### 3.2.1 Data collection and annotation

(1) Collection of dataset

In this study, a systematic multi-source data collection approach was applied to construct a comprehensive and diverse cotton disease image dataset by integrating web crawler technology, open datasets and institutional data. In the integration of the public dataset, we systematically integrate the PlantVillage cotton subset, AI agriculture track data, and Kaggle plant disease competition data, and collect a total of 4,078 images, which effectively improves the quality and consistency of the data. The details are as follows:

Open dataset (3,078 images, 75.4%):

PlantVillage Cotton Disease subset (1,532 images): 682 images of healthy leaves, 417 images of blight, and 321 images of leaf spot. Address: https://github.com/spMohanty/PlantVillage-Dataset

AI Challenger Ag Track Data (892 sheets): 283 sheets for leaf curl, 195 sheets for gray mold, and 214 sheets for yellow blight. Address:

https://tianchi.aliyun.com/competition/entrance/231702/information

Kaggle Plant Disease Competition data (654 sheets): 451 sheets for healthy leaves and 203 sheets for various diseases.

Address: https://www.kaggle.com/datasets/arjuntejaswi/plant-village

Web crawler to obtain data (total 1,000 sheets, 24.6%):

Targeted crawling using Scrapy framework for public images of agricultural research institutes Sources include:

Chinese Academy of Agricultural Sciences (CAAS) Digital Agriculture Platform: 428 images;

International Society for Plant Pathology (ISPP) public case library: 372 images;

Supplementary images of academic papers from agricultural colleges and universities: 200 images, as shown in Table 2:

**Table 2.** Distribution Map of Dataset Categories

| Disease Category | Label | Number of Samples | Percentage |
|---|---|---|---|
| Healthy leaves healthy | healthy | 1423 | 34.9% |
| Blight blight | blight | 782 | 19.2% |
| Leaf curl curl | curl | 612 | 15.0% |
| Grey mildew Grey mildew | Grey mildew | 486 | 11.9% |
| Leaf spot Leaf spot | Leaf spot | 459 | 11.3% |
| Yellow wilt wilt | wilt | 316 | 7.7% |

An example data set is shown in Figure 6. :

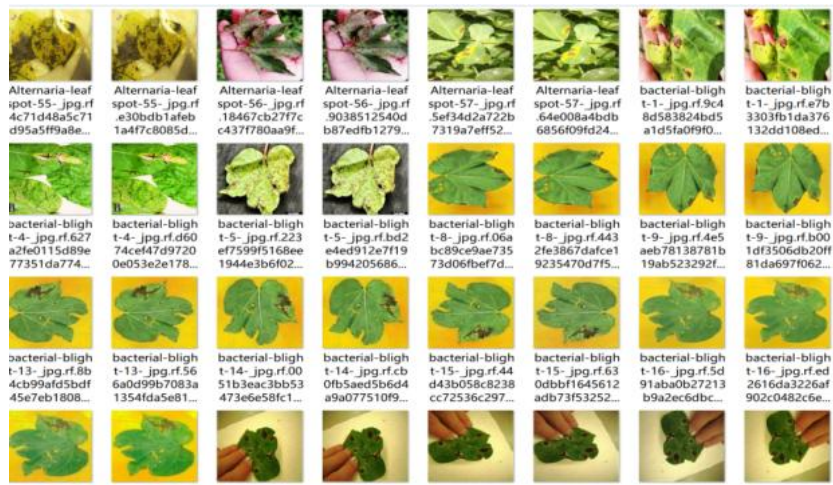

**Figure 6.** Example of the Cotton Disease Dataset

(2) Labeling of the dataset

In this study, LabelImg was used as the core annotation tool, and a set of systematic annotation process was used to carry out standardized labeling of the cotton disease image dataset, and the specific annotation process was strictly in accordance with the multi-level annotation specifications and the quality control system, and a complete set of disease category system was set up before the annotation work was started. classes.txt configuration file, we set up six categories of standard labels: ['blight', 'curl', 'grey mildew', 'healthy', 'leaf spot', 'wilt'], and established the corresponding mapping relationship between the Chinese and English labels, so as to ensure that the labeling process of the category selection is precise and error-free. As shown in Figure 7:

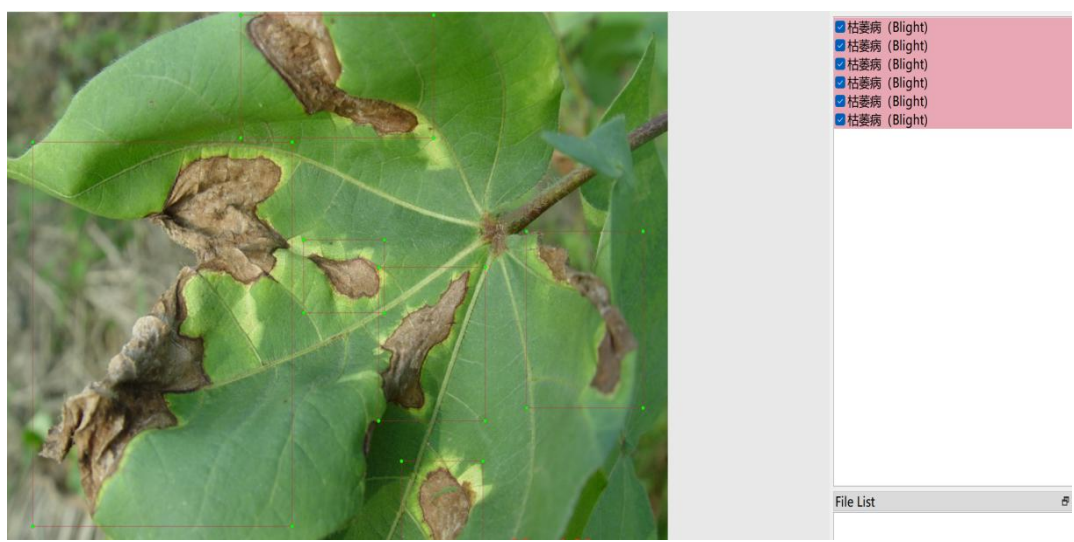

**Figure 7.** Label Instance

Validation of labeling consistency:

Using stratified random sampling, the number of samples ≥ 80 sheets from each of the six categories, while ensuring that the coverage of different collection sources, light conditions, and the stage of spot development, the IoU and category consistency as the main test criteria, if the IoU of the same target is greater than 0.85 and the category is consistent, then the labeling is considered to be consistent. The test results are shown in Table 3-3:

**Table 3.** Results of Annotation Consistency Verification

| Category | Sample size | Category concordance |
|---|---|---|
| blight | 83 | 92% |
| curl | 82 | 89% |
| Grew_mildew | 85 | 85% |
| healthy | 80 | 96% |
| Leaf_spot | 85 | 90% |
| wilt | 85 | 93% |

### 3.2.2 Data Preprocessing and Division

(1) Data Preprocessing

Size Normalization:With the help of bilinear interpolation algorithm, all input images are uniformly adjusted to the standard size of 640×640 pixels, and the edges are filled with grey background while maintaining the original aspect ratio, so as to reach the size uniformity without distorting the image. One.

Format Conversion:Unify the original RAW/JPEG format into JPG format to reduce the storage space occupied while preserving the image quality.

(2) Data set division

The study adopts a scientific data set division method to divide the labeled cotton disease image data set into training set, validation set and test set according to the proportion, and the specific division process and quality control are as follows:

Training set (80%): Adjust the model weights through the back-propagation algorithm to minimize the loss function, so as to enable the model to learn the key discriminative features of the cotton disease (e.g., lesion texture, color distribution, etc.) and to learn the key discriminative features of cotton disease (e.g., lesion texture, color distribution, etc.). spot texture, color distribution, etc.), and establish the mapping relationship between image pixels and disease categories to enhance the model robustness.

The validation set (10%) is used to guide the adjustment of parameters such as learning rate and batch size, monitor the validation loss to avoid overfitting, compare the actual performance of different network structures, and validate various data to improve the effectiveness of the method.

Test set (10%): provides real performance metrics of the model on unknown data, simulates the performance of real application scenarios, serves as the basis for fair comparison between different algorithms, and is the source of the final performance data for the report.

## 3.3 Model Training

### 3.3.1 Environment Installation

First install the VSCode compiler and the plug-in Remote-SSH, Chinese(Simplified), upload the prepared dataset to the cloud server and unpack it, download the corresponding code, create a new conda environment and activate it. Use the AliCloud image to install the environment, then download the weights, and finally present the following description of the environment is configured successfully

### 3.3.2 Creating and Modifying Files

(1) Create the train.py file

Create the train.py file under the path /root/lanyun-tmp/ultralytics/ and write it, load the model path to select the absolute path of yaml under models, and the data path under the training parameters to select the coco128.yaml absolute path

(2) Modify coco128.yaml file

Open coco128.yaml file, modify the path after train,val,test to the absolute path of your own dataset photo, and modify the label type to your own type.

## 3.4 Training results

In this study, yolov5,yolov8,yolov11 models were used for training respectively, and then compared the training results of several models, including the overall MAP value, and the MAP value of each subclass, etc., so as to select the optimal model, and its training results are shown in Figure 8., Figure 9, Figure 10.and Figure 11:

```
WARNING ⚠validating an untrained model YAML will result in 0 mAP.
Ultralytics 8.3.12 🚀 Python-3.10.16 torch-2.6.0+cu124 CUDA:0 (NVIDIA GeForce RTX 4090, 24210MiB)
YOLOv5 summary (fused): 211 layers, 2,182,834 parameters, 0 gradients, 5.8 GFLOPs
                 Class     Images  Instances      Box(P          R      mAP50  mAP50-95): 100%|         | 15/15 [00:
                   all        232        345        0.8      0.703      0.743      0.598
               healthy         86        154      0.831      0.773      0.833      0.616
                blight         45         59      0.895      0.724       0.79      0.684
                  curl         12         17      0.676      0.492      0.546      0.299
           Grey_mildew         27         27      0.898      0.978      0.982      0.982
             Leaf_spot         23         33      0.693      0.576      0.598      0.383
                  wilt         47         55      0.807      0.673      0.709      0.623
Speed: 0.1ms preprocess, 0.6ms inference, 0.0ms loss, 1.6ms postprocess per image
Results saved to runs/train/yolov52
```

**Figure 8 .**Training Result Diagram of Yolov5

The overall mAP50 was 0.743 and mAP50-95 was 0.598, indicating that the model has a strong target detection ability and can better accomplish the multi-category detection task of cotton diseases. The mAP50 of some categories with outstanding performance healthy, blight, and Grey_mildew reached 0.833, 0.790, and 0.982, respectively, and the mAP50-95 was also higher, indicating that the model learns the features of these categories sufficiently and has a high recognition accuracy. However, the mAP50-95 of curl and Leaf_spot are only 0.299 and 0.383, respectively, which are much lower than the other categories, indicating that the model is weak in distinguishing these categories.

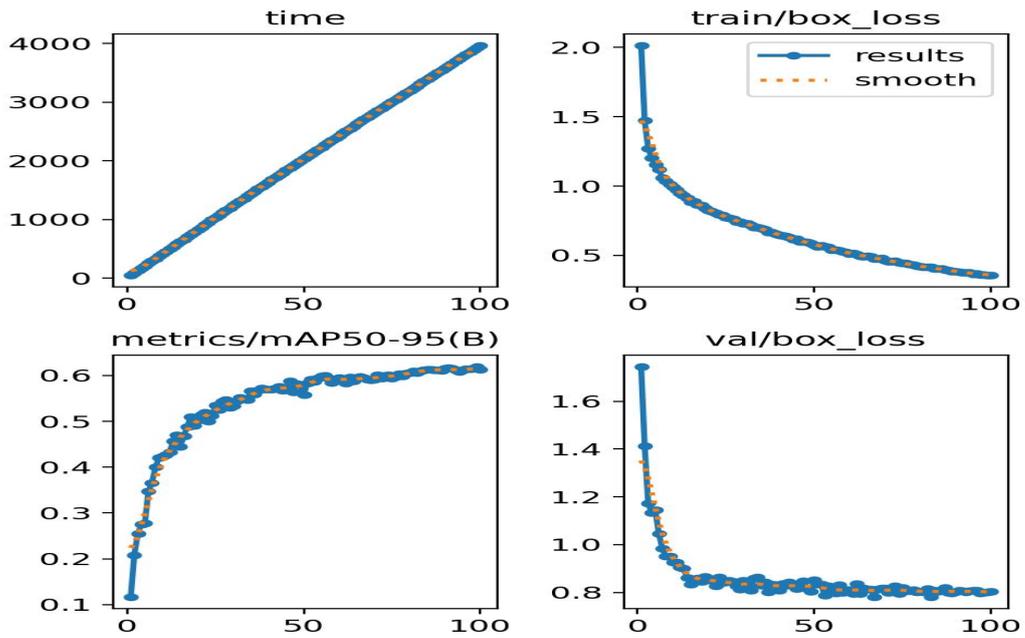

**Figure 9.** Training Result Diagram of Yolov8

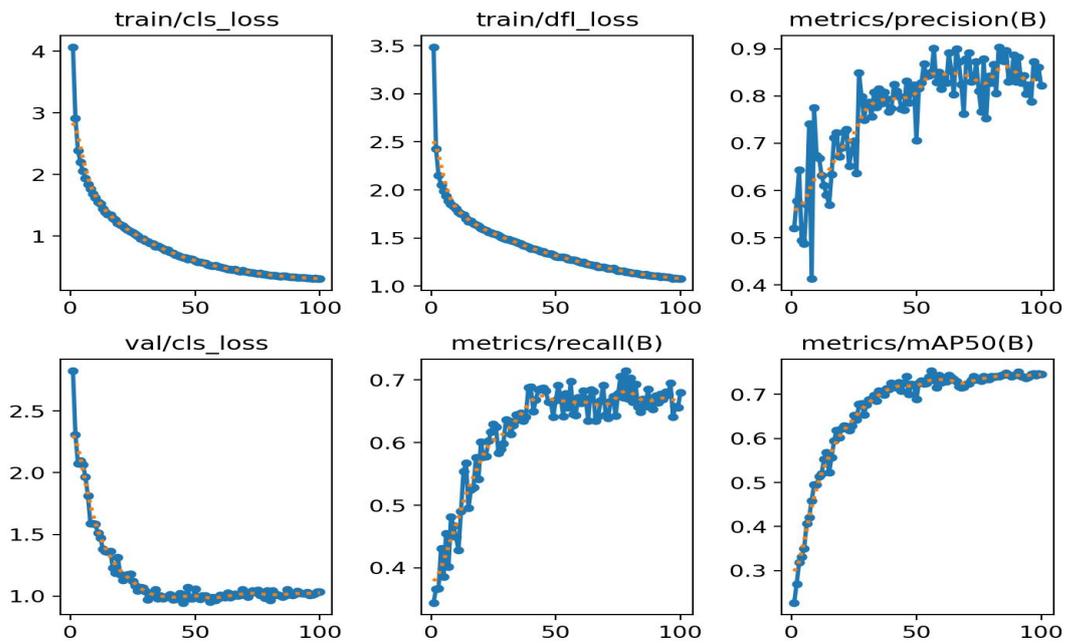

**Figure10.** Training Result Diagram of Yolov8

From the training results above, it can be seen that the training process of the Yolov8 model is relatively stable, the various loss functions and evaluation indexes perform well, the model can effectively learn the data features, and the core indexes such as precision rate, recall rate and mAP have reached a high level, indicating that the model has a strong target detection ability and generalization ability. There is no very prominent overfitting or underfitting phenomenon, and the trends of training and validation loss curves are consistent.

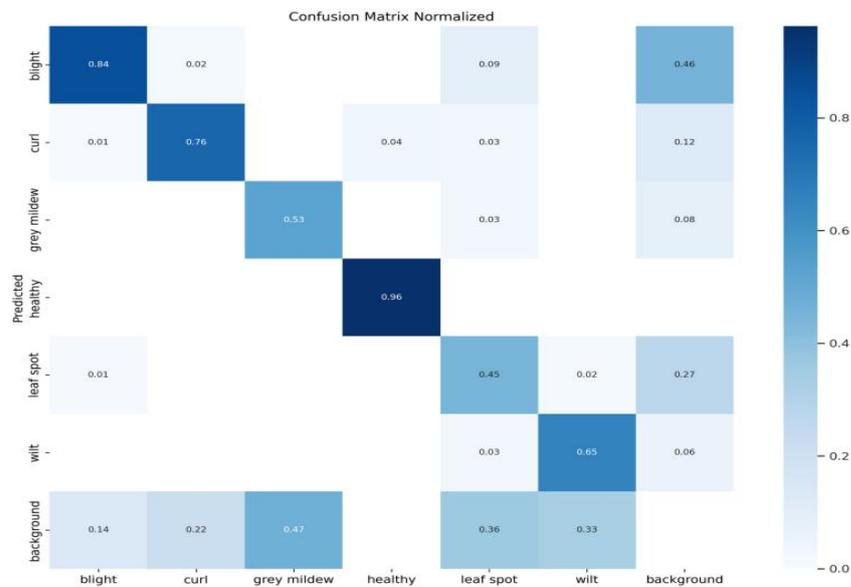

**Figure 11.** Training Result Diagram of Yolov11

The model recognizes most of the categories quite well, the training process is stable, the main diagonal is more obvious, the recognition accuracy of grey mildew, leaf spot and other categories is lower, and more confused with the background, it is suggested that the follow-up can rely on increasing samples, data enhancement or optimization of the model structure and other ways to improve the detection effect of these categories.

## 3.5 Comparison of result visualization

The visualization of this training process and results is realized with the help of Python's matplotlib library. matplotlib is a powerful data visualization tool, which can draw a variety of statistical graphs, such as line graphs, histograms, loss curves, confusion matrices and so on, and it has a wide range of applications in presenting the results of experiments on deep learning and machine learning, and it has a wide range of applications in presenting the results of experiments on deep learning and machine learning. matplotlib is used in this experiment. Matplotlib is utilized in this experiment for drawing and displaying. In this experiment, matplotlib is utilized to draw and display:

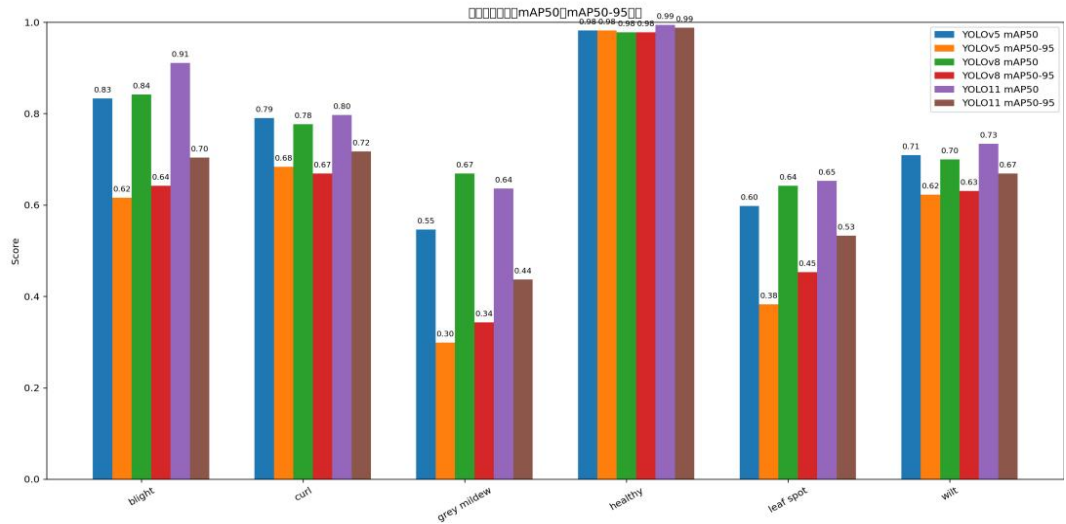

**Figure 12.** Comparison of Model Training Effects

Comparative analysis: This bar chart comparing mAP50 and mAP50-95 for each category of each model shows that YOLO11's mAP50 and mAP50-95 indexes are slightly higher than those of YOLOv5 and YOLOv8 in most of the categories, and the overall detection effect is the best. For example, on the categories of blight, curl, and healthy, YOLO11 has the highest mAP50 and mAP50-95, especially on the blight category where the mAP50 reaches 0.91, which is significantly better than the other models. However, the enhancement is not very significant overall. Whether it is mAP50 or mAP50-95, the gap between YOLO11 and YOLOv8 and YOLOv5 is relatively small, and the performance is very close between models in some categories such as HEALTHY, WILT, etc., and even in the categories of GREY MILDEW and LEFT SPOT, the advantage of YOLO11 is not very prominent.

In summary, YOLO11 outperforms YOLOv5 and YOLOv8 in the detection task of cotton disease categories, indicating that the model improvement has achieved some results, but the overall improvement is limited, the performance gap between models is not obvious, and further optimization can be done in the future to obtain a greater improvement.

# 4. Improvement of yolov11 model

## 4.1 Dynamically adjusting the ratio of Mosaic and MixUp

With the help of adjust_augmentation function to dynamically adjust the ratio of Mosaic and MixUp, the intensity of the data enhancement is gradually reduced with the advancement of the training, and for the small target categories, such as category 2 and category 4, the function adjust_mosaic_weights is used to dynamically adjust the ratio of Mosaic data enhancement. Mosaic_weights function is used to dynamically adjust the proportion of Mosaic data increase for small target categories, such as category 2 and category 4.

In the early stage of training, stronger data enhancements (e.g., Mosaic and MixUp) are used to increase data diversity and improve the generalization ability of the model. Gradually weaken the data enhancement strength in the later stages of training to help the model better converge to the real data distribution. Use a higher percentage of Mosaic data enhancement for small target categories to increase the training samples for small targets.

## 4.2 Dynamically adjust the weights according to the number of category instances

references the calculate_class_weights function, which dynamically calculates the category weights based on the number of instances of each category in the dataset, realizing automated weight calculation without manual adjustment.

Dynamically adjusting the weights according to the number of categories alleviates the problem of category imbalance, improves the detection of a few categories, prevents the model from over-biased towards the majority category, and improves the detection performance of a few categories. Automatically adjusting the weights based on the number of category instances reduces human intervention.

## 4.3 Introducing the C2PSA Module

The C2PSA module is introduced in the backbone of yolo11.yam to enhance the feature extraction ability of the model.The C2PSA module combines Channel Attention and Spatial Attention, which can effectively enhance the model's ability to focus on key features The C2PSA module combines Channel Attention and Spatial Attention to enhance the model's ability to focus on key features .

The introduction of C2PSA module can enhance the model's detection effect on complex scenes and small targets. Enhance the ability to adapt to complex scenes, in scenes with complex backgrounds or dense targets, the C2PSA module can effectively distinguish between targets and backgrounds. Enhance the overall detection performance and improve the mAP (mean average precision) of the model through more accurate feature extraction [14].

## 4.4 Adding Class Instance Statistics

Add class_counts field in coco128.yam file to record the number of instances of each class, which supports dynamic weight calculation and provides data support for calculate_class_weights function.

Add category instance statistics to alleviate the problem of category imbalance, relying on statistics on the number of category instances and dynamically adjusting the category weights to improve the detection effect of a few categories. Optimize the training process, provide category distribution statistics, facilitate analysis and optimize the data sampling strategy, adapt to different datasets, and work effectively on datasets with different category distributions.

## 4.5 Comparison and Analysis of Results

Map50 and Map50-95 are shown in Figure 13:

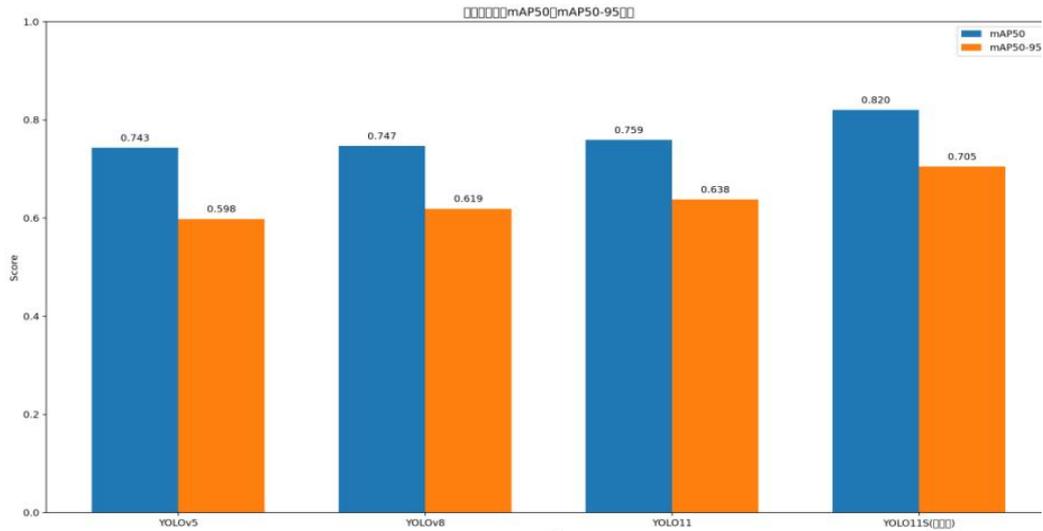

**Figure 13.** Comparison of the Effects of the Improved Yolov11

As can be seen from this overall mAP50 vs. mAP50-95 comparison histogram, the improved YOLO11S has made significant gains in both core metrics. Specifically, the mAP50 of YOLOv5, YOLOv8, and YOLO11 are 0.743, 0.747, and 0.759, respectively, while the mAP50 of YOLO11S (improved) improves to 0.820, a significant increase; in terms of mAP50-95, the mAP50-95 of YOLOv5, YOLOv8, and YOLO11 are 0.598, 0.619 and 0.638, while YOLO11S (improved) reaches 0.705, an even more prominent improvement.

This result indicates that the improved YOLO11S model outperforms the other comparative models in terms of overall detection accuracy [15], especially under the more stringent mAP50-95 metric, the advantage is more obvious. It indicates that the proposed improved method effectively enhances the target detection ability of the model.

The various sub-goals Map50 are shown in Figure 14.

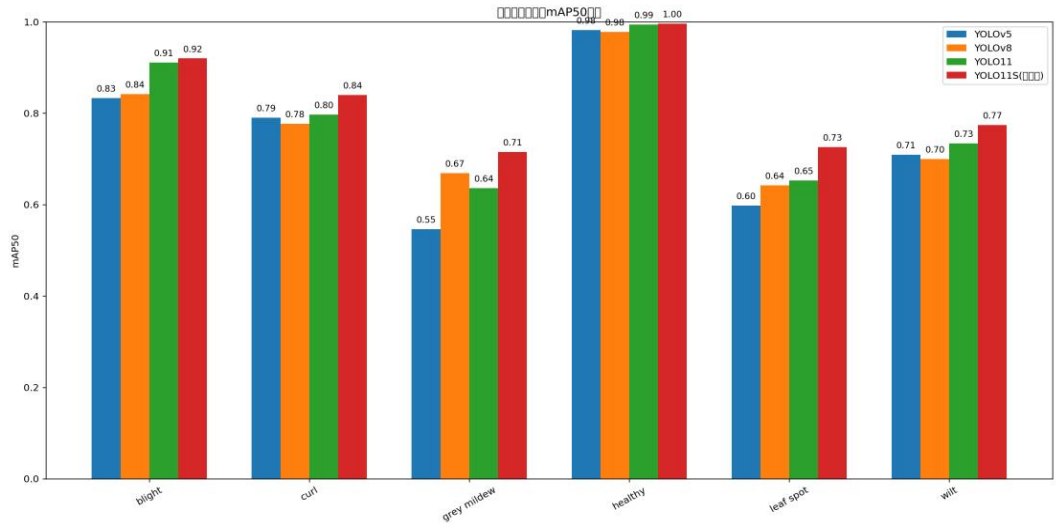

**Figure 14.** Comparison of the Effects of the Improved Yolov11 on Various Small Targets

As can be seen from the bar chart comparing the mAP50 of each subcategory, the improved YOLO11S achieves different degrees of improvement in all categories, and the mAP50 of the YOLO11S (improved) is higher than that of the YOLOv5, regardless of whether it is in the categories of blight, curl, grey mildew, healthy, leaf spot, or wilt, YOLOv8 and YOLO11, and the improvement is even more obvious in the grey mildew, leaf spot and wilt categories, which are originally more difficult to detect. Overall, YOLO11S (after improvement) maintains high accuracy in mainstream categories and makes breakthroughs in difficult-to-detect categories, indicating that the proposed improvement method can comprehensively enhance the detection ability of the model, allowing the model to be optimized in all indicators and have stronger practical application value.5 Implementation of Cotton Disease Detection System.

## 5. CONCLUSIONS

Focusing on the key issues of cotton disease detection in the process of intelligent development of agriculture, this study has carried out a series of innovative research work based on the YOLOv11 model, and achieved some key results. Specific research includes:

(1) Improvement of the accuracy of cotton disease detection in complex environments

Aiming at the problem of detecting cotton leaf diseases in the complex background of the field, the study improves the robustness of the existing model, and improves the recognition accuracy of multiple categories of diseases by virtue of optimizing the structure of the model and the training strategy, with special attention paid to the differentiation ability of small targets and overlapping targets.

(2) Research on the generalization ability of cotton disease models under small sample conditions

To address the problem of scarce and uneven distribution of disease samples in practical agricultural scenarios, efficient data utilization methods are explored. Research on how to improve the sample optimization and training strategy so that the model can still maintain high generalization performance under limited labeled data, and adapt to the disease detection needs of different cotton varieties and growth stages.

(3) Applied Research on Lightweight Model and Real-time Detection System

Facing the demand of real-time monitoring in farmland (e.g., UAVs, portable devices, etc.), research on how to reduce the computational complexity of the model under the premise of guaranteeing the detection accuracy. Design a lightweight solution applicable to low-power devices, and verify its real-time performance and stability in a real field environment.

(1) Data Enhancement Strategy Innovations

Generate richer training samples by designing or improving the data enhancement method, so as to improve the model's generalization ability and robustness. This innovation usually optimizes the training process of the model by adjusting the intensity, mode or dynamic changes of data enhancement.

(2) Category Imbalance Handling Innovation

Mitigates the negative impact of uneven distribution of categories in a dataset on model training by designing new methods or strategies. Such innovations usually improve the detection performance of a few categories by dynamically adjusting category weights, data sampling strategies, or loss functions.

(3) Model Structure Innovation

Enhance the feature extraction capability and task performance of the model by improving or designing new neural network structures. Such innovations usually involve the introduction of new modules, attention mechanisms, or optimization of network hierarchies to enhance the model's adaptability to complex scenarios.

(4) Dataset Configuration Optimization Innovations

Enhance the efficiency and effectiveness of model training by improving the organization of the dataset or adding statistical information. This kind of innovation usually supports the dynamic adjustment of training parameters by providing more detailed category distribution information or optimizing the data loading method.

In summary, in the field of theoretical innovation, the model's detection performance in complex agricultural scenarios is significantly improved by constructing a new network architecture and optimizing the training strategy, etc. The proposed dynamic feature fusion mechanism and intelligent sample allocation method effectively solve the technical problems such as insufficient detection capability of small targets and category imbalance. The use of hybrid precision training framework and model compression algorithm achieves the synergistic optimization of computational efficiency and precision.

In terms of technological innovation, the research successfully breaks through the limitations of traditional detection methods, and builds a complete technical system from algorithm design to engineering deployment. The designed lightweight detection scheme can support mobile deployment, providing a practical solution for real-time field monitoring, and the constructed standard dataset lays a key foundation for subsequent research.

In terms of optimization results, the cotton disease detection system constructed by the YOLOv11 model has improved its overall performance, but there are still areas that need to be improved. Although the accuracy of the model for common diseases has been improved with the introduction of dynamic feature fusion mechanism and other means, the detection of small-targeted diseases is still unsatisfactory, and there is still room for improvement, and the results of the test show that the current system can detect small spots in the early stages of the disease. The test results show that the current system has a relatively low recognition rate for early stage diseases with small lesions, which is mainly due to the insufficient feature extraction of tiny lesions in the complex background of the field.

## Acknowledgments

This research was funded by a grant from Key Laboratory for Crop Production and SmartAgriculture of Yunnan Province, Yunnan Provincial Agricultural Basic Research Joint Project (No.202301BD070001-203), Yunnan Provincial Basic Research Project (No. 202101AT070267), Yunnan Key Laboratory of Crop Production and Intelligent Agriculture 2024 Open Fund Project(No.2024ZHNY10) ,Yunnan Agricultural University Education and Teaching Reform Research Project(No.YNAUKCSZJG2023056),Yunnan Agricultural University University-level First-Class Undergraduate Curriculum Project(Big Data Storage and Processing Technology), the scientific research fund project of Yunnan Provincial Education Department (No. 2021J0943).

## Ethics and Consent to Participate declaration

Not applicable.

## References

[1] Jiali Chen. Research on plant disease recognition method based on deep migration learning[D]. Yangtze University,2024.DOI:10.26981/d.cnki.gjhsc.2024.000442.

[2] Yiyi Cao. Research on small sample cucumber disease recognition method based on multimodal fusion[D]. Anhui Jianzhu University2024.DOI:10.27784/d.cnki.gahjz.2024.000187.

[3] Liu Yechen. Research on small target detection algorithm for agricultural pests based on deep learning [D]. Hubei University,2024.DOI:10.27130/d.cnki.ghubu.2024.002343.

[4] Yu Cunjiang,Zhang Guangyu. Research on traffic sign detection method based on Faster R-CNN[J]. Information Recording Materials,2021,22(10):72-73.doi:10.16009/j.cnki.cn13-1295/tq.2021.10.033

[5] Zhenhan Zhang, Haixin Huang, Yi Wang, et al. Review of image segmentation technology based on deep learning[J]. Computer Application Abstracts,2024,40(16):158-160.

[6]Li Xiong. Design of vision-based casting localization system[D].Nanjing University of Science &Technology,2021.DOI:10.27241/d.cnki.gnjgu.2021.001172.

[7] Li Yuda,Wu Zhengping,Sun Shuifa,et al. Multi-species apple leaf disease detection with improved YOLOv5 algorithm[J]. Journal of Chinese Agricultural Mechanization,2024,45(12):230-237+F0003.DOI:10.13733/j.jcam.issn.2095-5553.2024.12.034.

[8] LU Jun,Xie Feng. Study on Detection Method of Rice Leaf Diseases Based on Improved YOLOv8 [J]. Journal of Guangxi Agriculture,2024,39(04):27-35+42.DOI:10.20160/j.cnki.ISSN1003-4374.2024.04.005.

[9] Zhou Xiushan,Wen Luting,Jie Baifei,et al.  Real-time Detection Algorithm of Expanded Feed Image on the Water Surface Based on Improved YOLOv11[J]. Smart Agriculture,2024,6(06):155-167.DOI:10.12133/j.smartag.SA202408014.

[10] Shi, Wu-Rui. Research on SAR ship detection combining supervised and semi-supervised learning[D].Guangxi University,2023.DOI:10.27034/d.cnki.ggxiu.2023.002223.

[11] Yang Chen. Research on semantic segmentation and target detection of remote sensing images based on deep learning [D].Ningxia University,2023.DOI:10.27257/d.cnki.gnxhc.2023.001154.

[12] Weiming Zhang, Caijuan Shi, Bijuan Ren, et al. Multi-scale Feature Pyramid Grid for Salient Object Detection[J]. Journal of Chinese Computer Systems,2022,43(05):1068-1074.DOI:10.20009/j.cnki.21-1106/TP.2020-1035.

[13] H.X. Zhao, M.Y. Shao, P.P. Pan, et al. A Training Dataset for Deep Neural Network Model Recognition of Common Cotton Diseases[J]. Journal of Agricultural Big Data,2023,5(04):47-55.DOI:10.19788/j.issn.2096-6369.230405.

[14] Fan B, Gao Weiwei, Shan Mingtao, et al. Lightweight semantic segmentation of UAV traffic scene objects combining attention mechanism and ghost feature mapping[J]. Journal of Electronic Measurement and Instrumentation,2023,37(03):21-28.DOI:10.13382/j.jemi.B2205751.

[15] Zhou Xuan,Ge Q,Shao WZ. Small Target Detection in UAV Aerial Images Based on High Resolution Feature Enhancement [J]. Journal of Data Acquisition & Processing,2024,39 doi:10.16337/j.1004-9037.2024.04.011

[16] Harshitha G, Kumar S, Rani S, et al. Cotton disease detection based on deep learning techniques[C]//4th Smart Cities Symposium (SCS 2021). IET, 2021, 2021: 496-501.doi: doi={10.1049/icp.2022.0393}

[17] Zekiwos M, Bruck A. Deep Learning-Based Image Processing for Cotton Leaf Disease and Pest Diagnosis[J]. Journal of Electrical & Computer Engineering, 2021.doi:10.1155/2021/9981437

[18] Manavalan R. Towards an intelligent approaches for cotton diseases detection: A review[J]. Computers and Electronics in Agriculture, 2022, 200: 107255.doi:10.1016/j.compag.2022.107255

[19] Latif M R, Khan M A, Javed M Y, et al. Cotton Leaf Diseases Recognition Using Deep Learning and Genetic Algorithm[J]. Computers, Materials & Continua, 2021, 69(3).doi=10.32604/cmc.2021.017364

[20] Faisal H M, Aqib M, Rehman S U, et al. Detection of cotton crops diseases u10.1038/s41598-025-94636-4sing customized deep learning model[J]. Scientific Reports, 2025, 15(1): 10766.doi:10.1038/s41598-025-94636-4

[21] Islam M M, Talukder M A, Sarker M R A, et al. A deep learning model for cotton disease prediction using fine-tuning with smart web application in agriculture[J]. Intelligent Systems with Applications, 2023, 20: 200278.doi=doi.org/10.1016/j.iswa.2023.200278

[22] Memon M S, Kumar P, Iqbal R. Meta deep learn leaf disease identification model for cotton crop[J]. Computers, 2022, 11(7): 102.doi=doi.org/10.3390/computers11070102

[23] Caldeira R F, Santiago W E, Teruel B. Identification of cotton leaf lesions using deep learning techniques[J]. Sensors, 2021, 21(9): 3169.doi=10.3390/s21093169

[24] Sarangdhar A A, Pawar V R. Machine learning regression technique for cotton leaf disease detection and controlling using IoT[C]//2017 international conference of electronics, communication and aerospace technology (ICECA). IEEE, 2017, 2: 449-454.doi=10.1109/ICECA.2017.8212855

[25] Gülmez B. A novel deep learning model with the Grey Wolf Optimization algorithm for cotton disease detection[J]. Journal of Universal Computer Science, 2023, 29(6): 595.doi=10.3897/jucs.94183

[26] Kumar R, Kumar A, Bhatia K, et al. Hybrid approach of cotton disease detection for enhanced crop health and yield[J]. IEEE Access, 2024.doi=10.1109/ACCESS.2024.3419906

[27] Hassan J, Malik K R, Irtaza G, et al. Disease Identification using Deep Learning in Agriculture: A Case Study of Cotton Plant[J]. VFAST Transactions on Software Engineering, 2022, 10(4): 104-115.

[28] Yang Z Y, **a W K, Chu H Q, et al. A comprehensive review of deep learning applications in cotton indust10.3390/plants14101481ry: From field monitoring to smart processing[J]. Plants, 2025, 14(10): 1481.doi:10.3390/plants14101481

[29] Devi C M, Vishva S P, Gopal M M. Cotton Leaf Disease Prediction and Diagnosis Using Deep Learning[C]//2024 International Conference on Advances in

Computing, Communication and Applied Informatics (ACCAI). IEEE, 2024: 1-7.doi:10.1109/ACCAI61061.2024.10602390

[30] Nazeer R, Ali S, Hu Z, et al. Detection of cotton leaf curl disease's susceptibility scale level based on deep learning[J]. Journal of Cloud Computing, 2024, 13(1): 50.doi:10.1186/s13677-023-00582-9

[31] Noon S K, Amjad M, Ali Qureshi M, et al. Computationally light deep learning framework to recognize cotton leaf diseases[J]. Journal of Intelligent & Fuzzy Systems, 2021, 40(6): 12383-12398.doi:10.3233/JIFS-210516

[32] Kinger S, Tagalpallewar A, George R R, et al. Deep learning based cotton leaf disease detection[C]//2022 International Conference on Trends in Quantum Computing and Emerging Business Technologies (TQCEBT). IEEE, 2022: 1-10.doi:10.1109/TQCEBT54229.2022.10041630

[33] Iqbal S, Ayaz A, Qabulio M, et al. A Deep Learning Based Model for the Classification of Cotton Crop Disease[J]. Technical Journal, 2024, 29(01): 61-68.url:https://tj.uettaxila.edu.pk/index.php/technical-journal/article/view/2139

[34] Tahir M S, Yaqoob A, Hamid H, et al. A methodology of customized dataset for cotton disease detection using deep learning algorithms[C]//2022 International Conference on Frontiers of Information Technology (FIT). IEEE, 2022: 77-81.doi:10.1109/FIT57066.2022.00024

[35] Dewangan U, Talwekar R H, Bera S. A systematic review on cotton plant disease detection & classification using machine & deep learning approach[C]//2023 1st DMIHER International Conference on Artificial Intelligence in Education and Industry 4.0 (IDICAIEI). IEEE, 2023, 1: 1-6. doi={10.1109/IDICAIEI58380.2023.10406941}

[36] Parashar N, Johri P. Deep learning for cotton leaf disease detection[C]//2024 2nd International Conference on Device Intelligence, Computing and Communication Technologies (DICCT). IEEE, 2024: 158-162.doi={10.1109/DICCT61038.2024.10533021}

[37] Dhage S, Garg V K. Cotton plant fungal disease classification using deep learning models[C]//2023 11th International Conference on Emerging Trends in Engineering & Technology-Signal and Information Processing (ICETET-SIP). IEEE, 2023: 1-5.doi={10.1109/ICETET-SIP58143.2023.10151608}

[38] Zhu D, Feng Q, Zhang J, et al. Cotton disease identification method based on pruning[J]. Frontiers in Plant Science, 2022, 13: 1038791.10.3389/fpls.2022.1038791

[39] Stephen A, Arumugam P, Arumugam C. An efficient deep learning with a big data-based cotton plant monitoring system[J]. International journal of information technology, 2024, 16(1): 145-151.doi:10.1007/s41870-023-01536-9

[40] Rahman M, Ullah M, Devnath R, et al. Cotton leaf disease detection: an integration of CBAM with deep learning approaches[J]. Int J Comput Appl, 2025, 975: 8887.doi:10.5120/ijca2025924487

[41] Singh G, Aggarwal R, Bhatnagar V, et al. Performance Evaluation of Cotton Leaf Disease Detection Using Deep Learning Models[C]//2024 International Conference on Computational Intelligence and Computing Applications (ICCICA). IEEE, 2024, 1: 193-197.doi={10.1109/ICCICA60014.2024.10584990}

[42] Tanwar P, Shah R, Shah J, et al. Cotton Price Prediction and Cotton Disease Detection Using Machine Learning[M]//Intelligent Data Communication Technologies and Internet of Things: Proceedings of ICICI 2021. Singapore: Springer Nature Singapore, 2022: 115-128.

[43] Bishshash P, Nirob A S, Shikder H, et al. A comprehensive cotton leaf disease dataset for enhanced detection and classification[J]. Data in Brief, 2024, 57: 110913.